\documentclass{article}
\usepackage[preprint]{neurips_2024}
\usepackage{amsmath}
\usepackage{stfloats}%
\usepackage[utf8]{inputenc} 
\usepackage[T1]{fontenc}    
\usepackage{hyperref}       
\usepackage{url}            
\usepackage{booktabs}       
\usepackage{amsfonts}       
\usepackage{nicefrac}       
\usepackage{microtype}      
\usepackage{xcolor}         
\usepackage{multirow}
\usepackage{chngcntr}
\usepackage{graphicx}
\usepackage{multirow}
\usepackage{xcolor}
\usepackage{colortbl}
\usepackage{makecell} 
\usepackage{amssymb}
\usepackage{bbding}
\usepackage{pifont}
\usepackage{wasysym}
\usepackage{utfsym}
\usepackage{fontawesome}
\usepackage{wrapfig}
\usepackage{float}
\usepackage{tabularx}

\definecolor{my_green}{RGB}{51,102,0}
\definecolor{my_red}{RGB}{204, 0, 0}
\definecolor{mygray}{rgb}{0.5, 0.5, 0.5} 

\definecolor{blue}{RGB}{71, 187, 222}
\definecolor{blue2}{RGB}{220, 234, 247}

\definecolor{ModelGreen}{RGB}{213,232,212}
\definecolor{ModelGrey_1}{RGB}{207,220,230}
\definecolor{ModelGrey_2}{RGB}{207,225,235}

\usepackage{booktabs} 
\usepackage{pifont}   
\usepackage{adjustbox} 
\usepackage{caption}  
\usepackage{amsmath}  
\usepackage{authblk}       
\usepackage{manyfoot}
\usepackage{subcaption}

\title{Video-XL-2: Towards Very Long-Video Understanding Through Task-Aware KV Sparsification}

\begin{document}

\author{
\textbf{Minghao Qin$^{1*}$  \ \ Xiangrui Liu$^{1,2*}$ \ \ Zhengyang Liang$^{1*}$  \ \ Yan Shu$^{1,3}$ \ \ Huaying Yuan$^{1,4}$ \ \ Juenjie Zhou$^{1,5}$ \ \ Shitao Xiao$^{1}$ \ \ Bo Zhao$^{1,2}$ \ \ Zheng Liu$^{1,6}$}

$^{1}$Beijing Academy of Artificial Intelligence  \; 
$^{2}$Shanghai Jiao Tong University \;
$^{3}$University of Trento \;
$^{4}$Renmin University of China \;
$^{5}$Beijing University of Posts and Telecommunications \;
$^{6}$Hong Kong Polytechnic University

Homepage: \url{https://unabletousegit.github.io/video-xl2.github.io/}}

\renewcommand{\thefootnote}{\fnsymbol{footnote}}
\footnotetext[1]{Equal contribution.}

\maketitle

\begin{abstract}
Multi-modal large language models (MLLMs) models have made significant progress in video understanding over the past few years. However, processing long video inputs remains a major challenge due to high memory and computational costs. This makes it difficult for current models to achieve both strong performance and high efficiency in long video understanding. To address this challenge, we propose Video-XL-2, a novel MLLM that delivers superior cost-effectiveness for long-video understanding based on \textbf{task-aware KV sparsification}. The proposed framework operates with two key steps: \textit{chunk-based pre-filling} and \textit{bi-level key-value decoding}. Chunk-based pre-filling divides the visual token sequence into chunks, applying full attention within each chunk and sparse attention across chunks. This significantly reduces computational and memory overhead. During decoding, bi-level key-value decoding selectively reloads either dense or sparse key-values for each chunk based on its relevance to the task. This approach further improves memory efficiency and enhances the model’s ability to capture fine-grained information.
Video-XL-2 achieves state-of-the-art performance on various long video understanding benchmarks, outperforming existing open-source lightweight models. It also demonstrates exceptional efficiency, capable of processing over 10,000 frames on a single NVIDIA A100 (80GB) GPU and thousands of frames in just a few seconds. Video-XL-2 has been made publicly available at \href{https://github.com/VectorSpaceLab/Video-XL}{\textbf{this repo}}. 
\end{abstract}
\section{Introduction}

Multi-modal large language models (MLLMs) have demonstrated remarkable progress in visual tasks that demand perception and reasoning abilities~\cite{reid2024gemini,gpt4o,Claude3, bai2025qwen2}. Recently, more and more research has extended MLLMs to video understanding~\cite{zhang2025videollama,li2024videochat,liu2025nvila,li2024llava,chen2025eagle}. given their significant potential in modeling temporal visual data. 
Existing MLLMs struggle with a major challenge in video understanding: the prohibitively high computational and memory costs, especially for long videos. Specifically, long videos consist of numerous frames, each producing a huge amount of visual tokens after visual encoded. As a result, the input often exceeds the model’s context window, making it computationally expensive and impractical for real-world applications. Even with extended context lengths via modifying model architecture or training strategy, processing such a large number of visual tokens remains inefficient and resource-intensive \cite{wang2024longllava, longva, chen2025eagle}. The primary solution to this problem is token reduction. Existing methods typically compress or downsample visual tokens per frame via an additional module~\cite{llama-vid2023,videollama,li2023videochat,li2024videochat,moviechat2023}. While these approaches certainly alleviate some memory and computational demands, they fundamentally struggle with the quadratic growth of computational FLOPs relative to the total number of input tokens. Consequently, for increasingly long video inputs, these models still face immense resource burdens and often incur critical information loss. Therefore, we propose Video-XL-2, a model featuring both advanced video understanding capabilities and outstanding efficiency, specifically designed for long video understanding.

$\bullet$ \textbf{Advanced Performance Construction}: Video-XL-2's advanced video understanding capabilities are fundamentally built upon two pillars: its innovative model architecture and an incremental training strategy. For the architecture, we adopt and refine the foundational structure of Video-XL-Pro~\cite{liu2025videoxlproreconstructivetokencompression}. Its main characteristic is an innovative module, Dynamic Token Synthesize (DTS), capable of effectively preserving rich visual information while mitigating redundancy across adjacent frames. Meanwhile, our incremental training strategy is designed to progressively robustly and comprehensively construct Video-XL-2's visual understanding abilities. This phased approach gradually equips the model with strong visual comprehension skills, ensuring its superior performance across diverse long video understanding tasks.

$\bullet$ \textbf{Comprehensive Efficiency Optimization}: Beyond its powerful understanding capabilities, Video-XL-2 integrates a comprehensive efficiency optimization strategy based on task-aware KV sparsification. Large Language Model (LLM) inference consists of two stages: pre-filling and decoding. In the pre-filling stage, Key-Value (KV) pairs corresponding to all input tokens are generated and stored in the KV cache. The decoding stage then iteratively generates output tokens one by one. Previous work~\cite{jiang2407minference,li2024snapkv,xiao2023efficient,zhang2023h2o,yuan2025native,lu2025moba} has observed significant KV sparsity in both pre-filling and decoding.
Based on this observation, we propose two main innovations, Chunk-based Pre-filling and Bi-level Key-Values (KVs) Decoding, to significantly alleviate the computational and memory burdens in both stages for LVU. During Chunk-based Pre-filling, we process the visual token sequence by dividing it into equal-length chunks. We then compute full attention within each chunk while applying sparse attention between chunks. This approach drastically reduces the overall computational and memory overhead during the pre-filling of vast token sequences. Subsequently, for the decoding stage, our Bi-level KVs Decoding strategy continues to manage the KV cache in chunks. The original KVs generated during pre-filling are designated as dense KVs. We then apply a downsampling operation on the dense KVs of each chunk to obtain sparse KVs. This process ensures that every chunk of the raw video input corresponds to a bi-level KV representation (both dense and sparse). During decoding, we selectively reload either dense or sparse KVs for each video chunk based on its relevance to the specific text query. This approach further optimizes memory usage and enhances the capture of fine-grained information.

Based on its innovative design, Video-XL-2 delivers significant advancements with the following key contributions:

\textbf{1. State-of-the-Art Performance}: Video-XL-2 achieve superior performance on various long video understanding benchmarks, outperforming all existing open-source lightweight (7B or 8B) MLLMs.

\textbf{2. Minimal Memory Footprint}: Our model exhibits exceptionally low memory requirements, enabling it to process up to 10,000 frames on a single 80G A100 GPU.

\textbf{3. Accelerated Inference Speed}: Video-XL-2 boasts high inference throughput, capable of processing thousands of frames significantly faster than other comparable open-source MLLMs.

\section{Related work}

\paragraph{Multimodal Large Language Models.}
The landscape of artificial intelligence has been reshaped by the emergence of Multimodal Large Language Models (MLLMs), which align powerful pre-trained vision encoders with large language models (LLMs) to perform complex vision-language tasks. Early pioneering works primarily focused on images. These models typically follow one of two dominant architectural paradigms. The first, popularized by models like LLaVA ~\cite{liu2023visual_llava} and MiniGPT-4~\cite{zhu2023minigpt4} , employs a simple projection module (e.g., a linear layer or a shallow MLP) to map visual features into the LLM's embedding space. The second, exemplified by Flamingo~\cite{alayrac2022flamingo} and BLIP-2~\cite{li2023blip2} , utilizes a more sophisticated cross-attention mechanism, often called a querying transformer (Q-Former), to distill visual information into a fixed number of learnable query tokens. The success of these image-based MLLMs naturally inspired their extension to the video domain \cite{videollama,li2023videochat,maaz2023videochatgpt,yang2025vidtext,li2025vid}. Initial works such as Video-LLaMA~\cite{videollama} and Video-ChatGPT~\cite{maaz2023videochatgpt} adapted these architectures by treating a video as an ordered sequence of frames. Visual features are extracted from each frame and concatenated, forming a long sequence of visual tokens that is subsequently processed by the LLM. While demonstrating promising video reasoning capabilities, this direct extension exacerbates a fundamental bottleneck: the quadratic complexity of the self-attention mechanism within the LLM. As video duration increases, the corresponding sequence of visual tokens grows linearly, leading to a surge in computational and memory costs. Furthermore, this approach naively treats all frames as equally important, failing to account for the significant temporal redundancy inherent in video data. This inefficiency motivates the development of more sophisticated methods designed specifically for long video inputs.

\paragraph{Long Video Understanding.}
To mitigate the challenges of processing long videos, the community has explored several research directions. We categorize these efforts into two main strategies: visual token reduction and attention sparsity. Visual Token Reduction strategies aim to decrease the number of visual tokens before they enter the main model, typically through compression, merging or downsampling. For instance, MovieChat~\cite{moviechat2023} and MA-LMM~\cite{malmm2024} manage information through memory banks that filter and store key details. Other methods, such as LLaMA-VID~\cite{li2024llama} and Video-CCAM~\cite{fei2024videoccam}, compress visual features into fixed query embeddings using cross-attention. Furthermore, VideoChat-Flash directly merges similar visual tokens via the ToMe~\cite{bolya2022token} approach, while VideoLLaMA2 employs spatial-temporal convolutions for effective visual token downsampling.
Inspired by advancements in long-context LLMs \cite{xiao2023efficient, zhang2023h2o, li2024snapkv,zhang2024long, wu2025lighter},  a second line of research leverages sparsity of self-attention mechanism to optimize computational cost for long video understanding. Previous researches ~\cite{chen2024image, he2024zipvl, cai2024pyramidkv} have observed high sparsity in attention computations within MLLMs for visual understanding, suggesting that full attention computation between all query and key vectors is often unnecessary. This observation is particularly evident in long video understanding. Building on this, MMInference~\cite{li2025mminference} proposed pre-defined sparse patterns to adaptively optimize computational FLOPs for different attention heads, achieving notable optimization during the pre-filling stage. 
Other approaches focus on Key-Value (KV) reduction to manage memory and computation. For instance, VideoChat-Flash~\cite{li2024videochat} attempts to accelerate pre-filling and decoding by eliminating KVs with low relevance to the query text in specific layers; however, its resulting FLOPs reduction is not substantial. Video-XL~\cite{shu2025video}  divides video input into chunks and streamingly processes video chunks to generate sparse KVs from special tokens inserted within visual tokens. These sparse KVs then attend to subsequent chunk encodings and the final decoding, significantly reducing computation overhead from extensive visual tokens. Similarly, ReTaKe~\cite{wang2024retake} also processes long videos at the chunk level, preserving full KVs for important selected frames within each chunk while dropping a fixed ratio of KVs for others based on attention scores. Despite these advancements, chunk-based approaches like \cite{shu2025video, wang2024retake} face a critical challenge: the number of KVs required to attend to the current chunk's encoding rapidly increases as more chunks are processed. This growing KV dependency across chunks makes it difficult to scale efficiently to extremely long videos, imposing a practical limit on achievable video length.

\section{Method}
\begin{figure*}[htbp] 
  \centering
  \includegraphics[width=0.90\textwidth]{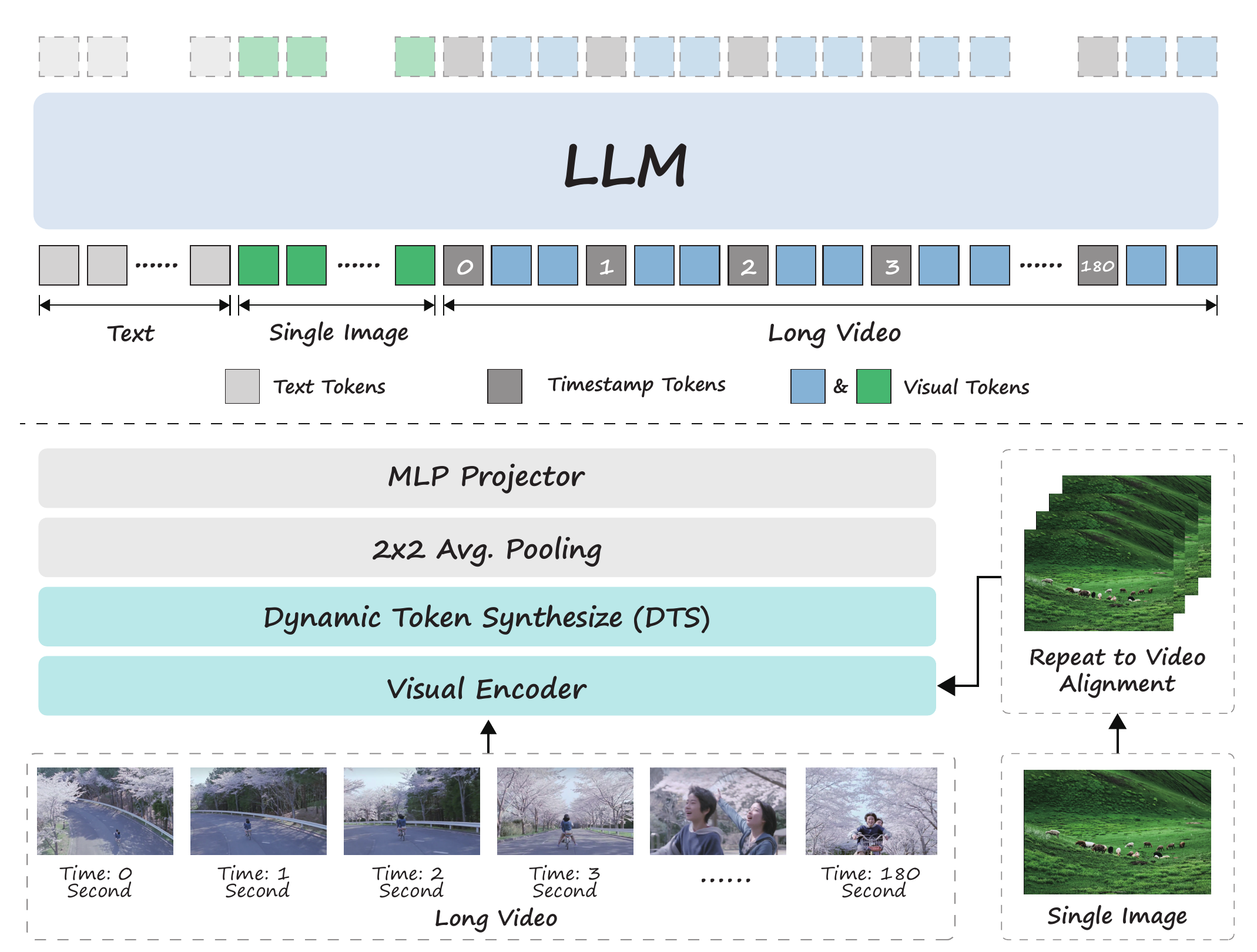}
  \caption{\textbf{The architecture of Video-XL-2.} The proposed Video-XL-2 comprises four main components: (1) Vision encoder to encode images and videos, (2) DTS to compress and make initial temporal modeling on visual features from vision encoder. (3) an MLP projector to project visual features into LLM embedding, and (4) a Large Language Model to process multi-modal inputs. Video-XL-2 interleaves timestamp tokens within the visual token sequence to enhance the model’s temporal awareness. Additionally, single image inputs are repeated four times to align with the video modality.}
  \label{fig:1}
  \vspace{-4mm}
\end{figure*}

\subsection{Model Architecture}
As shown in Figure ~\ref{fig:1}, the architecture of Video-XL-2 consists of four components: a vision encoder, dynamic token synthesis (DTS), an MLP projector, and a large language model (LLM). For the vision encoder, we adopt SigLIP \cite{tschannen2025siglip} to encode a visual input (i.e., single image or one frame) into dense visual features. The DTS module, positioned directly after the visual encoder, is constructed using a combination of spatio-temporal attention blocks and 3D convolutional layers. It is designed to process visual features extracted from four consecutive visual inputs (e.g., four video frames) as a single group. Specifically, while SigLIP independently encodes each frame in one group into dense visual representations, DTS then processes these grouped features to output a compact representation. This operation effectively compresses spatial-temporal redundancy while capturing dynamic motion patterns across the sequence. This design has been previously validated in Video-XL-Pro~\cite{liu2025videoxlproreconstructivetokencompression}. Following this, the architecture applies average pooling over adjacent features to further condense the representations. A two-layer MLP projector then processes these pooled features, projecting them into the LLM's embedding space. For our LLM, we adopt Qwen2.5-7B~\cite{qwen2025qwen25technicalreport}.

\subsection{Visual Input Processing}
How to process visual input before they were fed into LLM is a critical factor relevant to model performance.

\textbf{Frame Sampling Strategy.} For video input, we employ a simple yet more effective frame sampling strategy compared to standard uniform sampling or fixed 1 FPS sampling. The principle of this strategy is to prioritize oversampling as many frames as possible, up to a predefined frame upper bound. Specifically, the process unfolds in two steps: Initially, frames are sampled at 1 FPS from the raw video. Subsequently, the number of sampled frames is compared against this pre-defined limit. If the current frame count is below this upper bound, Video-XL-2 resamples at a higher rate (without exceeding the pre-defined maximum sampling rate) until the maximum frame count is reached.

\textbf{Temporal Information Injection.}
To enhance the model's temporal understanding capability, we prepend explicit timestamp tokens (e.g., Time: 4.0 Second) for every four consecutive frames group across the entire frames sequence. This direct timing information substantially improves the model’s temporal awareness. Complementing these explicit temporal signals, an implicit soft time embedding is also incorporated within the DTS module for each group of four frames. This additional embedding provides fine-grained temporal cues, enriching the local temporal context. 

\textbf{Image Modality Alignment.} For image input, each image is repeated four times to create a static, video-like sequence. This process effectively aligns the image modality with the video input format. This simple yet effective design facilitates the transfer of visual understanding capabilities learned from static images to dynamic video contexts, enabling a unified processing pipeline.

\begin{table*}[h]
\small 
\centering
\begin{tabularx}{\linewidth}{>{\centering\arraybackslash}p{2cm}>{\centering\arraybackslash}X >{\centering\arraybackslash}X >{\centering\arraybackslash}X >{\centering\arraybackslash}X} 
\toprule
\textbf{Stage} & \textbf{Stage 1} & \textbf{Stage 2}& \textbf{Stage 3} & \textbf{Stage 4}\\
\midrule
\textbf{Objective} & Initialize the weight of DTS module. & Initialize the weight of MLP projetor. & Building the foundation for visual understanding. & Enable the model to handle a diverse range of visual tasks. \\
\midrule
\textbf{Trained Modules} & DTS Module & MLP Projector & Full Params & Full Params \\
\midrule
\textbf{Training Data} & 250K image-cap. 750K video-cap. & 2M image-cap. 400K video-cap. & 5M image-cap. 2.7M video-cap. & 3M image-inst 2.5M video-inst \\
\bottomrule
\end{tabularx}
\caption{The overview of each training stage for the Video-XL-2}
\vspace{-4mm}
\label{tab:training_stages}
\end{table*}

\subsection{Training Datas and Strategy}
Inspired by previous studies~\cite{li2024videochat, zhang2025videollama, longvila, liu2025nvila}, we designed an incremental training strategy to equip our models with robust visual understanding capabilities step by step, as Table~\ref{tab:training_stages} shows. This incrementality manifests in two key aspects: the selection of training data and tunable modules. We divided the entire training process into four stages. The initial two stages are dedicated to initializing the DTS and MLP projector, respectively, establishing a solid foundation for subsequent training. For these early stages, smaller-scale image and short video caption datasets suffice. In the latter stages, we activate all model parameters for training and utilize larger-scale, higher-quality image and video datasets include both caption and various instruction tuning data to cultivate strong visual understanding capabilities.

\textbf{Stage-1: DTS Pre-training.}
This initial stage aims to initialize the DTS module's weights for subsequent training. During this process, the MLP projector and LLM are detached, while the vision encoder and DTS are jointly trained through an efficient reconstruction training objective, as proposed by Video-XL-Pro~\cite{liu2025videoxlproreconstructivetokencompression}. We utilize a dataset of 250k images-caption and 750k short video-caption for this stage, enabling DTS to effectively process both static image and dynamic video inputs. Notably, the vision encoder is frozen throughout this stage, with only the DTS module being updated.

\textbf{Stage-2: Video-Language Alignment.}
Similar to Stage-1, this stage focuses on initializing the MLP projector. During this stage, only the MLP projector was trained. The MLP projector serves as a cross-modal aligner, responsible for aligning the compact visual representations generated by the vision module (including DTS) with the text representation space of the LLM. For this stage, we utilize 2 million pairs of small-scale image-caption and 400k short video-caption data. Pre-training the MLP projector in this manner significantly benefits the convergence of subsequent training stages.

\textbf{Stage 3: Visual Pre-training.}
Building foundational visual understanding is the primary objective of this stage. We train the model with all parameters active using a substantial dataset comprising 5 million image-caption pairs and 2.7 million short video-caption pairs. This broad exposure to diverse visual-caption data is crucial for establishing the model's basic yet robust capability to interpret visual information. This stage, therefore, equips the model with the necessary elementary visual comprehension capability before proceeding to more complex tasks.

\textbf{Stage 4: Image and Video Instruction Tuning.}
This final stage focuses on instruction tuning to enable Video-XL-2 to handle a wide variety of downstream tasks. For this, we collected 5.5 million instruction fine-tuning samples, comprising 3 million image-based and 2.5 million video-based examples. To ensure the model retains its capacity for fine-grained understanding while significantly expanding its comprehension of long videos, we mix both short and long video data with image data. All model parameters are active during this stage. Upon completion, our goal is to obtain a versatile base model for comprehensive video understanding, facilitating subsequent optimizations and diverse applications.

\subsection{Efficiency Optimization}
The aforementioned model and training designs equip Video-XL-2 with robust visual understanding capabilities, yielding promising performance for long video tasks. However, efficiency remains an equally critical dimension for practical long video understanding. To address this challenge, we capitalize on the KV sparsification observed in long video understanding to propose a comprehensive optimization strategy targeting both the pre-filling and decoding phases of inference.

\begin{figure}[h]
    \centering
    \includegraphics[width=1.0\linewidth]{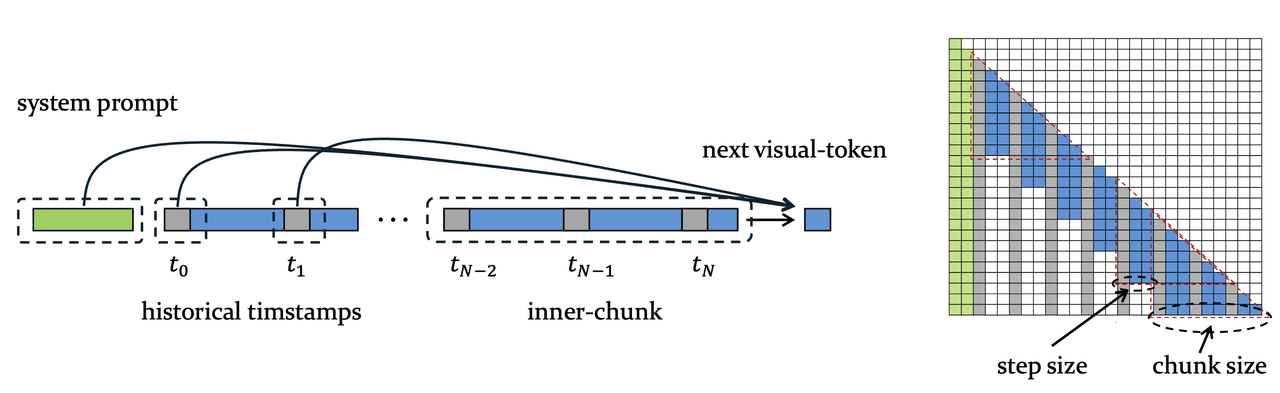}
    \caption{\textbf{Chunk-based Pre-filling Illustration.} In chunk-based pre-filling, the current processing chunk only attend to itself, historical timestamp tokens and the system prompt, as depicted in the left subfigure.
    And the right subfigure illustrates the current chunks for processing are decided by a sliding chunk window.}
    \label{fig:pre-filling}
\end{figure}

\textbf{Chunk-based Pre-filling.} The computational complexity of attention operations scales quadratically with the length of the input context. For long videos, this leads to significant and often unacceptable memory and computational overhead, making pre-filling the primary efficiency bottleneck. Drawing upon the observation that attention within VLMs exhibits notable sparsity—implying that much of the full attention computation for visual tokens is redundant—we introduce an innovative chunk-based pre-filling approach. As the Figure~\ref{fig:pre-filling} shown, Instead of computing attention across the entire visual token sequence, our method processes it in chunks. A straightforward and naive approach would be to assume the semantic information of each chunk is entirely independent. This implies that when computing attention, each chunk only needs to attend to itself, thereby achieving extremely low attention computation. However, the associations between chunks are, in fact, not negligible. Consider this point, when processing the current chunk, we allow it to attend not only to its own tokens but also to historical timestamp tokens from preceding chunks. The KVs of historical timestamp tokens provide a abstract, coarse-grained historical information without incurring extensive memory cost. Furthermore, we employ a sliding chunk window with a defined step size to determine the current chunk for processing. This mechanism allows the current chunk to attend to overlapping portion of the preceding chunk's detailed visual information. Through this approach, Video-XL-2 encodes long video input chunk by chunk. Each forward primarily involves computing attention for the visual tokens within a single chunk and a minimal set of historical timestamp token KVs. This design significantly reduces memory consumption and substantially accelerates the pre-filling stage, effectively mitigating the quadratic scaling challenge for long video inputs.

\begin{figure}[h]
    \centering
    \includegraphics[width=0.8\linewidth]{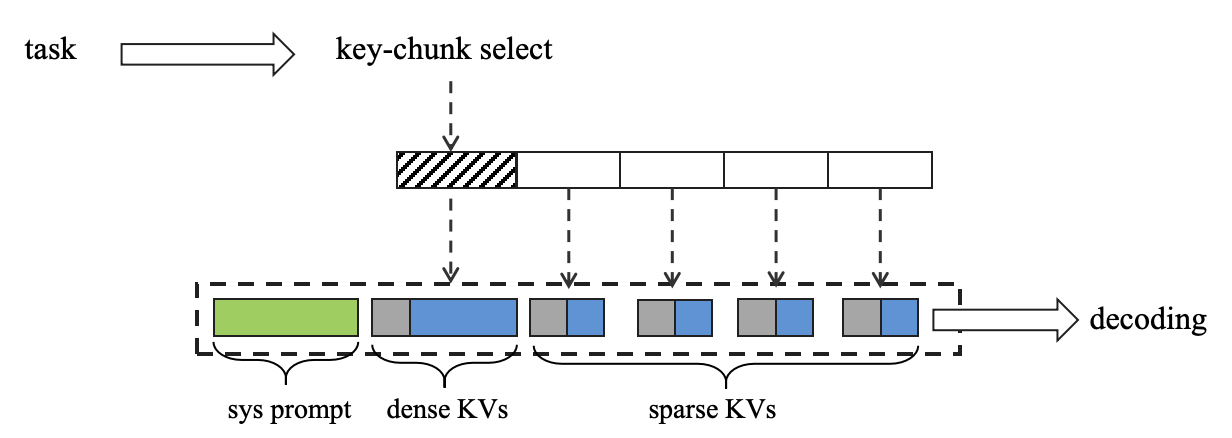}
    \caption{\textbf{Bi-level KVs decoding.} Bi-level KVs comprise both dense KVs (derived from the full video input) and sparse KVs, where the latter are obtained by downsampling the former at a chunk level. During decoding, it selectively reloads either dense KVs for video chunks highly relevant to the specific task query text, or sparse KVs for less relevant chunks, optimizing memory while preserving critical information.}
    \label{fig:decoding}
\end{figure}

\textbf{Bi-level KVs Decoding.} 
Building upon efficient pre-filling, we further optimize decoding for long videos by proposing a novel Bi-level Key-Value (KV) Decoding approach as the Figure~\ref{fig:decoding} shown. This approach reduces the KV cache size during the decoding phase, thereby increasing generation speed. After the KVs corresponding to the entire video input are generated (which we define as dense KVs), we proceed to chunkify these dense KVs. For each resulting KV chunk, we then apply a pooling operation to generate corresponding sparse KVs. This process yields bi-level KVs (dense and sparse) in chunks. Crucially, the same chunking strategy is applied to the raw video input, establishing a direct mapping between each video input chunk and its associated bi-level KVs. Both these dense and sparse KVs are then stored in an offline memory. When a specific task query text is provided, this approach first computes a relevance score for each video input chunk using a relevance oracle. This oracle can be implemented via various methods such as a Multimodal Large Language Model (MLLM) embedder~\cite{zhu2023languagebind, wang2024internvideo2, zhou2024megapairs}, attention score mechanism or a more complex, custom-designed relevance computation pipeline~\cite{yuan2025memory}. Based on these relevance scores, our Bi-level KVs Decoding strategy then selectively loads KVs for each video chunk. Specifically, it loads dense KVs for several video chunks that are highly relevant to the text query, thereby preserving fine-grained details. Conversely, for most other less relevant chunks, it loads sparse KVs, which retain only abstract or coarse-grained information to provide global background information. These reloaded KVs are then concatenated to form a new, complete "mixed KV" set. This design makes the entire KV cache lighter and more effective, significantly improving both efficiency and performance, especially for tasks requiring detailed perception within long video contexts.

\begin{table*}[h]
\centering
\addtolength\tabcolsep{-2.4pt} 
\resizebox{1\linewidth}{!}{
\begin{tabular}{lcc|c|c|c|c|c|c|c|c}
\toprule
\multicolumn{1}{c}{\multirow{2}{*}{Model}} & 

\multicolumn{1}{c}{Visual Input} & \multicolumn{1}{c|}{Flops} & \multicolumn{1}{c|}{MLVU Dev}  & \multicolumn{1}{c|}{MLVU Test} & \multicolumn{1}{c|}{VideoMME}& \multicolumn{1}{c|}{\multirow{2}{*}{LongVideo.}} & \multicolumn{1}{c|}{\multirow{2}{*}{LVBench}} & \multicolumn{1}{c|}{\multirow{2}{*}{VideoEval-Pro}} & \multicolumn{1}{c|}{\multirow{2}{*}{CharadesSTA}} & \multicolumn{1}{c}{\multirow{2}{*}{V-STaR}} \\

\multicolumn{1}{c}{} & \multicolumn{1}{c}{Len. (k)} & \multicolumn{1}{c|}{(G)} & M-avg & M-avg & w/o sub & \multicolumn{1}{c|}{} & \multicolumn{1}{c|}{} & \multicolumn{1}{c|}{} & \multicolumn{1}{c|}{} & \multicolumn{1}{c}{} \\ 
\midrule
\rowcolor{gray!15}\multicolumn{11}{c}{\textbf{Closed-source Models}} \\

GPT-4o  & - & - & {64.6} & \textbf{54.9} & 71.9 & \underline{66.7} & 64.4 & \textbf{34.2} & - & -  \\
Gemini-2.5-Pro & - & -  & \underline{81.2} &  -  & \textbf{87.0}& - & \textbf{69.2}  & - & - & - \\
Seed1.5-VL-8B & - & - & \textbf{82.1} & {-} & \underline{77.9}& \textbf{74.4}  & \underline{64.6} & - & \underline{64.7} & {-} \\
Eagle2.5-8B & 131.1 k & 8.8×\(10^3\) & 77.6 & {-} & 72.4 & 66.4 & {-} & {-} & \textbf{65.9} & {-} \\

\midrule
\rowcolor{gray!15}\multicolumn{11}{c}{\textbf{Open-source Large Models}} \\ 

InternVL2.5-78B & 16.4 k & 3.0×\(10^3\) & 75.7 & {-} & 72.1& 63.6  & {-} & {-} & {-} & {-} \\
Qwen-2.5-VL-72B & 24.6 k & 5.1×\(10^3\) & 74.6 & {-} & 73.3 & 60.7 & {47.3}  & {-} & 50.9 & {-} \\
LLaVA-Video-72B & 43.3 k & 11.1×\(10^3\) & 74.4 & {-} & 70.6 & 61.9 & {-} & {-}& {-} & {-} \\

\rowcolor{gray!15}\multicolumn{11}{c}{\textbf{Open-source Light Models}} \\ 

InternVL2.5-8B & 16.4 k & 339.7 & {68.9} & {-} & 64.2 & 60.0  & {-} & {24.6} & {-} & {7.8} \\
Qwen-2.5-VL-8B & 24.6 k & 590.4 & 70.2 & {-} & 65.1 & 56.0 & {45.3} & \underline{27.7} & 43.6 & {7.6}  \\
LLaVA-Video-8B & 43.3 k & 1.4×\(10^3\) & 70.8 & {-} & 63.3 & 58.2  & {-} & {24.2} & {-} & {12.2}  \\

VideoChat-Flash-8B & 8.2 k & {142.8}  & \underline{74.6} & {-} & 65.3 & \textbf{64.7} & \underline{48.2} & {27.0} & {48.0} & {-}  \\
NVILA-8B & 8.2 k & 142.8 & 70.1 & {-} & 64.2 & 57.7 & {-} & {-} & {-} & {-}   \\
VideoLLaMA3-8B & 17.3 k & 364.5 & 73.0 & \underline{47.7} & \underline{66.2} & 59.8 & 45.3 & {-} & \textbf{60.7} & \textbf{23.1} \\

\rowcolor{blue2}\textbf{Video-XL2-8B} & 14.9 k & \textbf{142.0} & \textbf{74.8} & \textbf{52.2} & \textbf{66.6} & \underline{61.0} & \textbf{48.4} & \textbf{28.6} & \underline{54.2} & \underline{21.3} \\ 
\bottomrule
\end{tabular}}
\caption{Results on long video understanding and temporal grounding benchmarks. All results for Video-XL-2 are obtained with Chunk-based Pre-filling and Bi-level KVs Decoding, enabled. ``LongVideo.'' is an abbreviation for LongVideoBench.} 

\vspace{-4mm}
\label{tab:main_1} 
\end{table*}

\section{Results}

\subsection{{Benchmarks and Metrics}}
We empirically evaluate the effectiveness of Video-XL-2 based on several popular long video understanding benchmarks. (1). MLVU~\cite{zhou2024mlvu}, a comprehensive benchmark that consists of multiple choice and generation tasks. (2). Video-MME~\cite{videomme}, another extensive benchmark covering videos of diverse genres and lengths (short, medium, and long). (3). LongVideoBench~\cite{wu2024longvideobench}, a benchmark designed for tasks that require precise retrieval and reasoning over detailed multi-modal information within extended inputs. (4). LVBench~\cite{wang2024lvbenchextremelongvideo}, extreme longvideo understanding. (5). VideoEval-Pro~\cite{ma2025videoeval}, a realistic LVU benchmark, is specifically designed with open-ended short-answer questions that demand comprehensive understanding of the full video. We further evaluate the model’s timestamp awareness using the Charades-STA~\cite{huang2024vtimellmcharadessta} temporal grounding dataset and V-STaR~\cite{cheng2025vstarbenchmarkingvideollmsvideo} long video temporal grounding dataset. To specifically highlight our efficiency advantages, we use ``Visual Input Length (k)'' and ``FLOPs (G)'' as key metrics in ``Main Results'' section. Both metrics are calculated based on the average of the maximum frame counts utilized across all long video understanding benchmarks.

\subsection{Main Results}
We present the performance of Video-XL-2 (enable Chunk-based Pre-filling and Bi-level KVs Decoding) on popular LVU and temporal grounding benchmarks in Table \ref{tab:main_1}. Notably, it outperforms mainstream open-source methods on the MLVU dev and test sets. Its dev set performance even surpasses GPT-4o and well-known open-source models like VideoChat-Flash-8B and NVILA-8B, which are similar in scale to it, demonstrating highly competitive results. For the general-video-understanding benchmark VideoMME, Video-XL-2 achieves 66.6\% accuracy under the "no subtitles" settings, reaching sota levels compared to other models. It outperforms InternVL2.5-8B and Qwen-2.5-VL-8B.
It also achieves highly competitive results on the LongVideoBench and LVBench, two benchmarks specifically designed for long video evaluation (ranking second and first, respectively). Video-XL-2's leading performance on VideoEval-Pro's realistic open-ended question-answering tasks further confirms its robust capabilities.

In the general temporal understanding benchmark Charades-STA and the long video temporal understanding benchmark V-STaR, Video-XL-2 achieves advanced levels, indicating our method's strong temporal grounding capabilities. Finally, while being highly competitive, our method has the lowest Flops of all, achieving the best balance between efficiency and performance.

\subsection{Efficiency Analysis}
This section demonstrates the significant efficiency gains achieved through our comprehensive strategies: chunk-based pre-filling and bi-level KV decoding strategies. The efficiency metrics used are FLOPs (\%) and KV Cache (Decoding)(\%). The latter quantifies the percentage of KV cache used during the decoding phase. As detailed in Table \ref{tab:efficiency}, when Video-XL-2 is equipped with chunk-based pre-filling, it reduces average FLOPs to just 48.8\% of the original, with only a minimal performance drop. Specifically, the average performance decline is less than 0.5 across the MLVU, VideoMME, and LongVideoBench benchmarks. Furthermore, our bi-level KV decoding strategy substantially reduces KV cache usage during the inference stage. We've achieved an average KV cache reduction of 38.8\% across MLVU Dev, VideoMME, and LongVideoBench, which notably compensates minimal performance drop from chunk-based pre-filling. 

\begin{figure}[h]
    \centering
    \begin{subfigure}{0.47\linewidth}
        \centering
        \includegraphics[width=\linewidth]{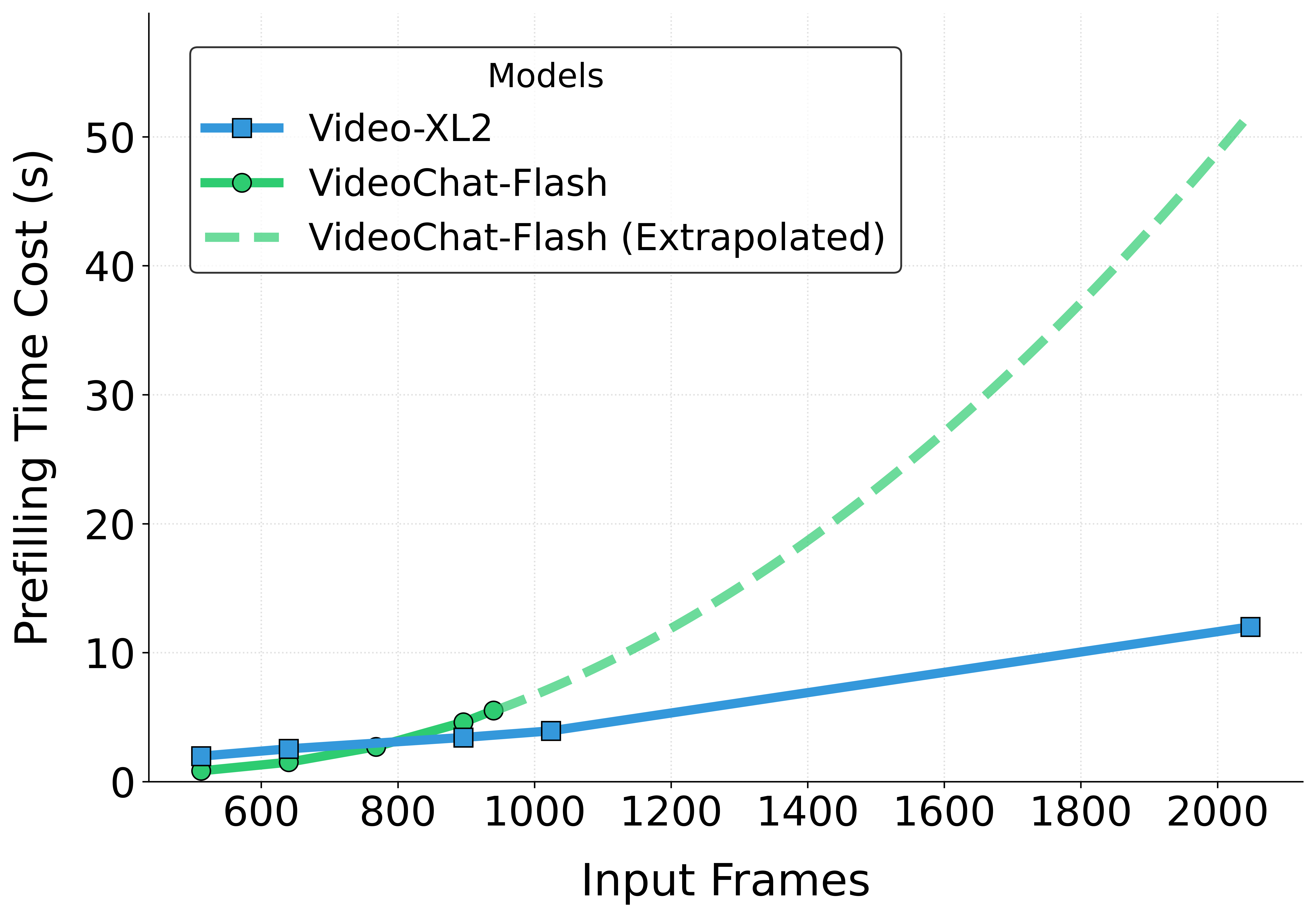}
        \caption{Speed Performance}
        \label{fig:speed}
    \end{subfigure}
    \hfill
    \begin{subfigure}{0.47\linewidth}
        \centering
        \includegraphics[width=\linewidth]{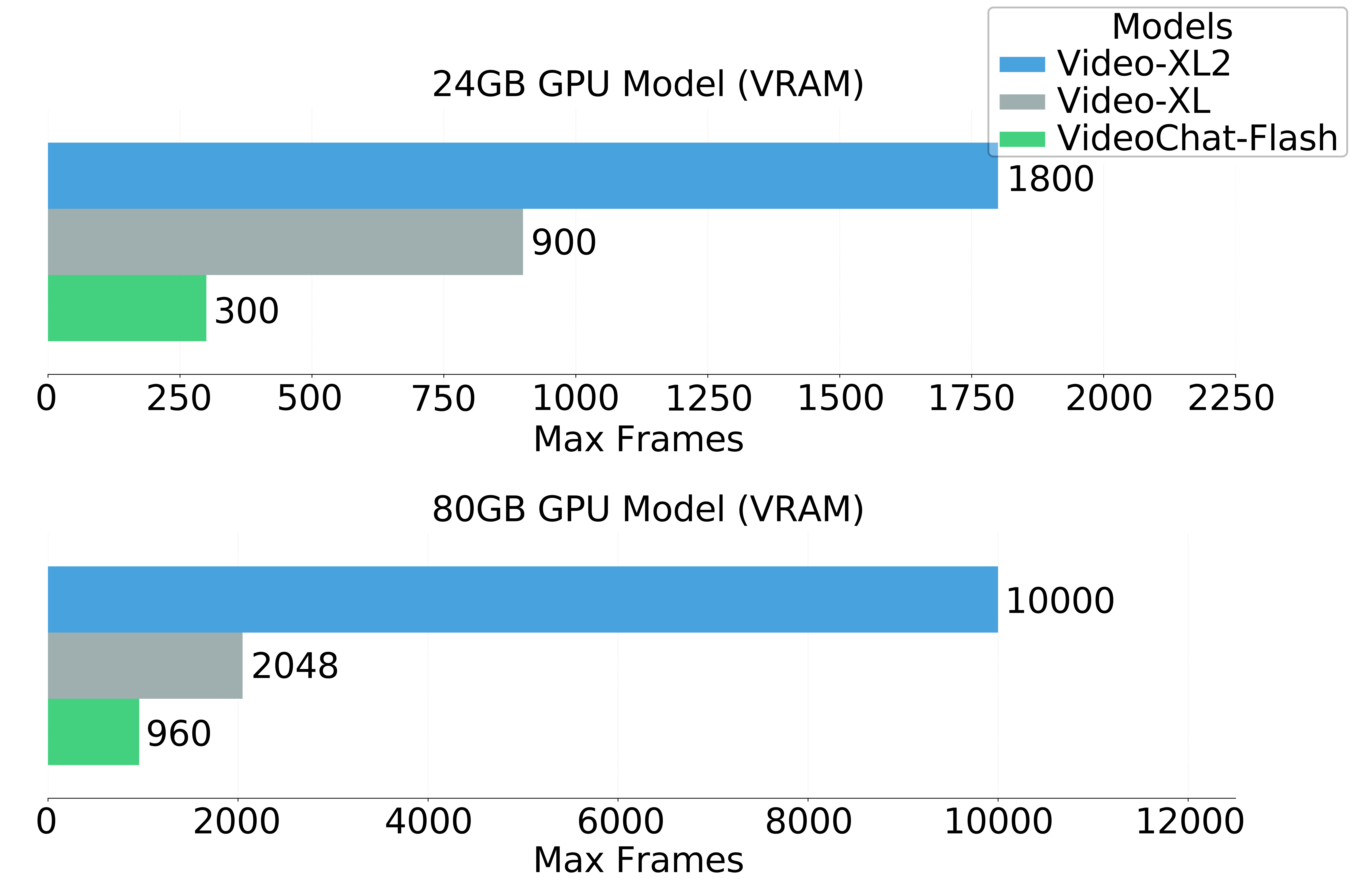}
        \caption{Memory Usage}
        \label{fig:memory}
    \end{subfigure}
    \caption{Efficiency Analysis Illustration. All results measured using eager attention for fair comparison.}
    \label{fig:both}
\end{figure}

Beyond these static metrics, we also conducted experiments to analyze speed and memory usage with varying input frame counts, with all results measured using eager attention for fair comparison. As Figure ~\ref{fig:speed} shows, the pre-filling time cost scales almost linearly with the number of input frames. This linearity strongly indicates that chunk-based pre-filling is highly flexible and scalable for even longer video inputs. Additionally, Figure ~\ref{fig:memory} demonstrates the clear advantage of Video-XL-2 in memory efficiency when both strategies are combined. Our model can process thousands of frames and up to 10,000 frames on GPUs with 24GB and 80GB of memory, respectively. In conclusion, our comprehensive efficiency strategy, which combines chunk-based pre-filling with bi-level KV decoding, proves to be both highly effective and remarkably efficient, making Video-XL-2 a practical solution for long video understanding.

\begin{table*}[h]
\centering
\adjustbox{max width=\linewidth}{
\begin{tabular}{l c c c c c}
\toprule


\multicolumn{1}{l}{\multirow{2}{*}{\textbf{Model}}}
& \multicolumn{1}{c}{\textbf{Avg. FLOPs}} & \multicolumn{1}{c}{\textbf{Avg. KV Cache}}& \multicolumn{1}{c}{\textbf{MLVU}} & \multicolumn{1}{c}{\textbf{VideoMME}} &
\multicolumn{1}{c}{\multirow{2}{*}{\textbf{LongVideo.}}}
\\
\textbf{} & \textbf{(\%)} & \textbf{(Decoding) (\%)} & \textbf{Dev Set} & \textbf{w/o subtitles} & \textbf{} \\

\midrule
Video-XL2 & 100\% & 100\% & 74.0 & 66.3 & 60.0 \\
+Chunk-based Prefilling & 48.8\% & 100\% & 73.5 & 66.0 & 59.7 \\
+Bi-level KVs Decoding & 48.2\% & 61.2\% & 74.8 & 66.6 & 61.0 \\
\bottomrule
\end{tabular}
}
\caption{Effect of two efficient strategies.}
\label{tab:efficiency}
\end{table*}

\subsection{Extra-long Video Scenario}
\begin{figure}
    \centering
    \includegraphics[width=1.0\linewidth]{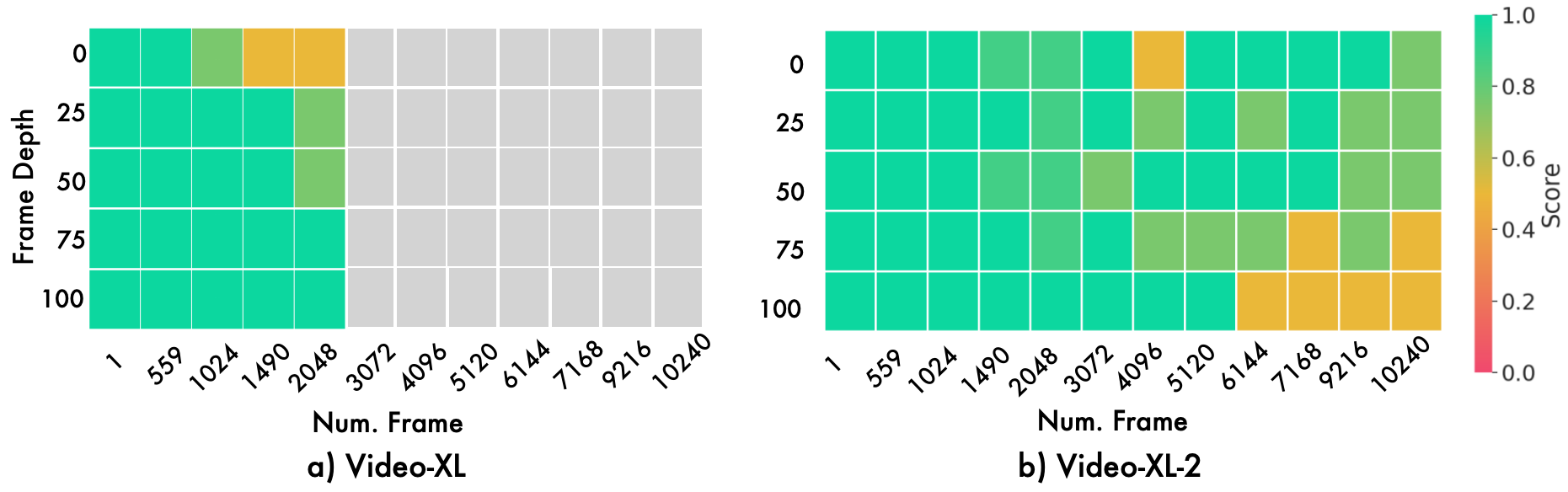}
    \caption{Needle in Haystack Evaluation.}
    \label{fig:needle}
\end{figure}
Processing extra-long video inputs presents a significant challenge for VLMs, as it demands models that are both memory-efficient for long sequences and capable of precisely capturing key detailed information within extended contexts. Thanks to its comprehensive efficiency strategy, Video-XL-2 can effortlessly process up to 10,000 frames on a single GPU while precisely capturing detailed information in these lengthy inputs. To evaluate this, we utilized question-answer pairs from \cite{wei2025videorope} and videos exceeding one hour from the VideoMME as our ``haystack" videos to deliver a ``Needle in Haystack'' Evaluation. 
As depicted in Figure~\ref{fig:needle}, Video-XL could only process 2048 frames. While Video-XL-2 could handle up to 10,000 frames and achieved strong performance on this challenging task.

\section{Conclusion}
In this paper, we presented Video-XL-2, our latest light vision-language model demonstrating strong capabilities in long video understanding and temporal grounding. We demonstrated that Video-XL-2 achieves state-of-the-art results on most benchmarks for long video understanding and exhibits competitive performance on temporal grounding tasks, when compared to other light open-source VLMs. Beyond its strong benchmark performance, Video-XL-2 also showcases remarkable inference efficiency. This is primarily attributed to our two innovative technical contributions: the chunk-based pre-filling and the bi-level KVs decoding. These advancements position Video-XL-2 as a highly capable and practical solution for addressing the challenges of processing and comprehending long video content. Future work will explore extending Video-XL-2's capabilities to even longer video durations and investigating its application in broader scenarios.
\clearpage
\small 
\bibliographystyle{unsrt} 
\bibliography{main}

\end{document}